\documentclass[nohyperref]{article}

\usepackage{microtype}
\usepackage{graphicx}
\usepackage{subfigure}
\usepackage{booktabs} % for professional tables
\usepackage{multicol}
\usepackage{caption}
\usepackage{arydshln}

\usepackage{hyperref}

\usepackage[accepted]{icml2022}

\usepackage{amsmath}
\usepackage{amssymb}
\usepackage{mathtools}
\usepackage{amsthm}

\usepackage[capitalize,noabbrev]{cleveref}

\theoremstyle{plain}

\theoremstyle{definition}

\theoremstyle{remark}

\usepackage[textsize=tiny]{todonotes}

\icmltitlerunning{SanitAIs: Unsupervised Data Augmentation to Sanitize Trojaned Neural Networks}

\begin{document}

\twocolumn[
\icmltitle{SanitAIs: Unsupervised Data Augmentation to \\
           Sanitize Trojaned Neural Networks}



\icmlsetsymbol{equal}{*}

\begin{icmlauthorlist}
\icmlauthor{Kiran Karra}{equal,yyy}
\icmlauthor{Chace Ashcraft}{equal,yyy}
\icmlauthor{Cash Costello}{yyy}
\end{icmlauthorlist}

\icmlaffiliation{yyy}{Research and Exploratory Development Department, Johns Hopkins University//Applied Physics Laboratory, Laurel, MD, USA}

\icmlcorrespondingauthor{Kiran Karra}{kiran.karra@jhuapl.edu}

\icmlkeywords{Machine Learning, ICML}

\vskip 0.3in
]



\printAffiliationsAndNotice{\icmlEqualContribution} % otherwise use the standard text.

\begin{abstract}
Self-supervised learning (SSL) methods have resulted in broad improvements to neural network performance by leveraging large, untapped collections of unlabeled data to learn generalized underlying structure. In this work, we harness unsupervised data augmentation (UDA), an SSL technique, to mitigate backdoor or Trojan attacks on deep neural networks. We show that UDA is more effective at removing trojans than current state-of-the-art methods for both feature space and point triggers, over a range of model architectures, trojans, and data quantities provided for trojan removal. These results demonstrate that UDA is both an effective and practical approach to mitigating the effects of backdoors on neural networks.
\end{abstract}

\section{Introduction}
\label{sed:intro}

Deep neural networks (DNNs) continue to achieve state-of-the-art performance on a wide variety of tasks.  This has led to additional research investigating their robustness, trustworthiness, and reliability including vulnerabilities to adversarial attacks.  Trojan attacks, also called backdoor or trapdoor attacks, are a training time adversarial attack.  These attacks modify a machine learning model through some algorithmic procedure to respond to a specific trigger in the model's input. When this trigger is present, the model will infer a pre-programmed response that could have potentially malicious consequences in a deployed setting.

Using the standard nomenclature, we define a \textit{trigger} as a model-recognizable characteristic of the input data that is used by an attacker to insert a trojan, and a \textit{trojan} to be the alternate behavior of the model when exposed to the trigger, as desired by the attacker.
Trojan attacks are effective if the triggers are rare or impossible in the normal operating environment, so that they are not activated in normal operations and do not reduce the model's performance on normal inputs. Additionally, the trigger is most useful if it can be deliberately activated at will by the adversary in the model's operating environment, either naturally or synthetically.  

A trojan attack can be implemented by manipulating both the training data and its associated labels \cite{gu2017badnets}, directly altering a model’s structure \cite{zou2018potrojan}, or adding training data that have correct labels but are specially-crafted to produce the trojan behavior \cite{turner2018clean}.  Perhaps the easiest way to poison a neural network with a trojan is by manipulating the training data through data poisoning. It has been shown that minuscule amounts of modified data are needed to insert the trojan behavior~\cite{dai2019textpoison}.
%Trojan attacks’ specificity differentiates them from the more general category of “data poisoning attacks”, whereby an adversary manipulates a model’s training data to make it ineffective.  

However, detecting poisoning in the data seems impractical due to the enormous size of datasets required to train state-of-the-art deep learning models. Even if datasets are controlled, trojans can be embedded into models in a continual learning environment by drifting trusted data away from the expected distribution~\cite{kantchelian2013approaches, zhang2020online}. %, and then the models are distributed to unsuspecting users.
Instead of analyzing the training data which may not even be available for some models, a common approach is detecting the trojan in the model. At the time of this writing, the Intelligence Advanced Research Projects Activity (IARPA) is holding a competition, called TrojAI, on detection of trojans in neural networks~\cite{iarpa_2019}. 
After the trojan is detected, one may sanitize the model through a sanitization algorithm, if one is known, or simply discard it.
Our proposed trojan mitigation strategy is to bypass the need for detection and develop a process which effectively cleanses a model of trojans if they are present, but has minimal effect on the model's performance for its intended task. In this case, the process produces an new model where triggers are rendered ineffective while preserving accuracy on non-triggered data. 

Mitigation approaches in the literature include NeuralCleanse \cite{wang2019neural}, fine pruning \cite{liu2018fine}, bridge mode connectivity \cite{zhao2020bridging}, and neural attention distillation \cite{li2021neural}.  In this work, we propose a self-supervised method that uses unsupervised data augmentation (UDA) \cite{xie2019unsupervised} and empirically show that it is more effective at mitigating various types of triggers than previously published state-of-the-art methods. Strengths of this UDA-based approach include: 1) not having to select hyperparameters which are difficult to chose in real-world scenarios, and 2) using unlabeled datasets to further boost performance.  In this paper, we begin by summarizing existing approaches to trojan mitigation and discuss respective limitations.  We then describe UDA and explain how to apply it to trojan mitigation.  Next, we present our experimental setup and results, and finally conclude with a discussion of our approach's advantages while including suggestions for future work.

\section{Current Approaches for Trojan Mitigation}
\label{sed:cur_app}
Research into trojan mitigation has existed since the introduction of trojans in DNNs \cite{gu2017badnets}. An early approach was fine-tuning, which involves further training of the trojaned DNN on a smaller, vetted dataset \cite{liu2017neural}. This approach can require a significant amount of labelled data before it is effective. 
Fine-pruning~\cite{liu2018fine}, a combination of pruning and fine-tuning, was another early mitigation technique but its performance can be sensitive to training hyperparameters. %, and we expect final performance on clean data will always be less than or equal to the performance of the original model. 
NeuralCleanse \cite{wang2019neural} went in a different direction using gradient information to reverse engineer the trigger before mitigating its effects. Reverse engineering the trigger is computationally expensive and error-prone especially when considering global triggers like image filters.  

\begin{figure}[t]
    \centering
    \includegraphics[width=0.65\columnwidth]{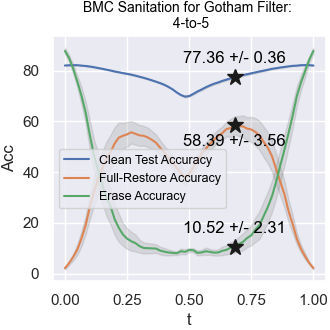}
    \caption{Accuracy of models along the bridge mode, trained with a 5\% subset of CIFAR-10 data.  The models were configured to classify the label ``4'' as the label ``5'', when an Instagram Gotham filter is present. Here, the $x$ axis represents $t$, the value along the curve of equivalent loss which was learned by the BMC algorithm, and the $y$ axis represents the accuracy. The star indicates the optimal value of $t$, which maximizes the clean data accuracy while minimizing the effect of the trigger.}
    \label{fig:bmc_gotham_0.05}
\end{figure}

A recent method is Bridge Mode Connectivity (BMC), which relies on a geometric discovery that a curve of equivalent loss exists between two models, and that this curve, parametrized by a single hyperparameter $t$, can be discovered through standard optimization techniques \cite{garipov2018loss}.  BMC was shown to be useful in mitigating trojans in \citet{zhao2020bridging}. By choosing a model along the curve of equivalent loss between two triggered models sufficiently far away from the originals (curve endpoints), but not being so far away from the end points that performance degrades to unacceptable levels, the trojan's effect could be removed without significant loss of performance on clean data.  Fig.~\ref{fig:bmc_gotham_0.05} shows this concept more concretely.  To use BMC for trigger removal, one first learns the curve of equivalent loss, and then chooses a value of $t$ (which corresponds to a different set of model weights along that curve with the same loss) that satisfies operational requirements for clean data accuracy versus triggered data accuracy.

While shown to be effective, two practical considerations make BMC infeasible for realistic trojan removal:

\begin{enumerate}
    \item For BMC to be effective, the hyperparameter $t$ which indicates the model to be chosen along the loss surface, needs to be chosen correctly.  This is not possible without access to triggered data.  This is demonstrated in Fig.~\ref{fig:bmc_gotham_0.05} where the optimal value of $t$ is defined as the point that maximizes the clean data accuracy while minimizing the trigger's effect and is indicated by the star.  It is seen that varying $t$ by even $10\%$ will drastically change both the clean data accuracy and the triggered data accuracy.
    \item Another concern is that BMC requires two models to operate, and it is unlikely that a second, similar-quality model will be readily available.  A possible workaround is to fine-tune an initial model to generate a second model, which can then be used for the bridge connection.  While practically possible, this is unsatisfying because the same data used to fine-tune will be used to build the bridge-mode, and the data processing inequality specifies that information cannot be gained by further processing.
\end{enumerate}

Neural Attention Distillation (NAD) is another recent approach for trojan mitigation \cite{li2021neural}.  NAD works by fine-tuning the trojaned model with an additional loss term derived from a ``teacher'' model and an attention operator. The attention operator is applied to blocks of convolution layers of both the teacher model and the trojaned model, and the loss term is setup to minimize the difference between the attention values of the two models. The teacher model should ideally be one that is not influenced by the trigger in the trojaned model. In practice, however, the teacher is the trojaned model fine-tuned on available clean data and \citet{li2021neural} shows that using the fine-tuned model can be effective under the assumed operating conditions. Currently, this approach is limited to networks with convolutional layers.

%A more practical consideration is that in a realistic use-case, it will be difficult to find clean data that is an exact subset of the original data which was used to trigger the network.  --> I'm commenting this out, b/c even though this is the point we want to make, most of our results w/ UDA are shown, using some % of teh original training data.  We do need to show results w/ out the original training data also, to prove our point, but, it doesn't seem central to what we're claiming currently.

\section{UDA-based Trojan Mitigation}
\label{sec:uda}
From previous experiments, we know that having more supervised data results in better mitigation performance. However, getting large amounts of data for cleaning neural networks is often not feasible due to the costs of data curation and annotation. Thus, our primary motivation for this work is to develop a trojan mitigation technique that uses more easily obtainable unlabeled data. Self-supervised learning (SSL) is a method of training DNNs that does not require labels and has been shown to increase performance in many areas of deep learning.  Many variants of self-supervised algorithms exist in the literature, but in this work we focus on unsupervised data augmentation (UDA).

UDA is an SSL technique which attempts to teach models to learn underlying structure in data, thereby  increasing model robustness and performance~\cite{xie2019unsupervised}.  The structure of the data is learned through the UDA objective (Eq.~\ref{eq:uda_objective}), which adds an unsupervised consistency loss $\mathcal{J_{\text{unsup}}}(\theta)$ to the original supervised loss $\mathcal{J_{\text{sup}}}(\theta)$.

\begin{equation}
    \mathop{\text{min}}_{\theta} J(\theta) = \mathcal{J_{\text{sup}}}(\theta) + \mathcal{J_{\text{unsup}}}(\theta)
    \label{eq:uda_objective}
\end{equation}

The unsupervised consistency loss (Eq.~\ref{eq:uda_unsup}) measures the difference in the consistency of predictions made by the DNN between unsupervised data points and random perturbations of those same unsupervised data points. Minimizing Eq.~\ref{eq:uda_unsup} results in maximizing this consistency.

%\begin{equation}
%    \mathcal{J_{\text{sup}}}(\theta) = \mathbb{E}_{x \sim p_L(x)} \left[ -\textbf{log}p_{\theta}(y | x)  \right]
%    \label{eq:uda_sup}
%\end{equation}

\begin{equation}
    \mathcal{J_{\text{unsup}}}(\theta) = \lambda \mathbb{E}_{x \sim p_U(x)} \mathbb{E}_{\hat{x} \sim q(\hat{x}|x)} \left[ \textbf{CE}\left( p_{\Tilde{\theta}}(y|x) || p_{\theta}(y | \hat{x}) \right) \right]
    \label{eq:uda_unsup}
\end{equation}

In (\ref{eq:uda_unsup}), $x$ is the input, the output distribution is given by $p_\theta(y|x)$, $\textbf{CE}$ denotes cross entropy, $q(\hat{x}|x)$ is the data augmentation transformation.   $\Tilde{\theta}$ is a fixed copy of the current parameters $\theta$ indicating that the gradient is not propagated through $\Tilde{\theta}$, and $\mathcal{D}(\cdot || \cdot) $ indicates computation of divergence between the two distributional arguments.  

UDA was created to increase DNN performance when there is a limited amount of supervised training data. The algorithm was shown to be successful in both image and text domains across a wide range of network architectures. Simultaneously, researchers in adversarial machine learning have discovered that enforcing consistency in model predictions is important, primarily under the popular inference-style adversarial attacks~\cite{cohen2019certified}. At the time of this writing, we are unaware of any approaches applying these ideas to the trojan problem. 

\begin{figure}[ht!]
    \centering
    \includegraphics[scale=0.4]{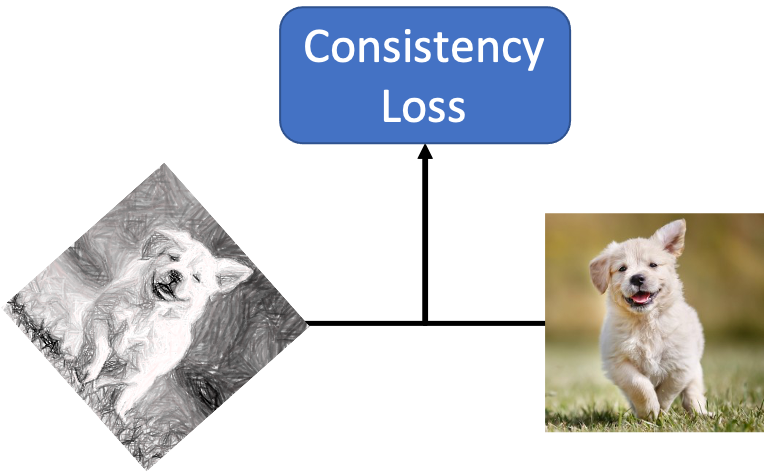}
    \caption{Example of consistency training}
    \label{fig:uda_intuition}
\end{figure}

Intuition for why consistency loss can be helpful in data poisoning is shown in Fig.~\ref{fig:uda_intuition}. Here an image and a rotated and style modified version of that image are shown as inputs to a consistency loss function. The consistency loss encourages the network to make the same prediction regardless of perturbation. In the data poisoning domain, triggers are designed to be highly specific, to avoid being activated arbitrarily~\cite{karra2020trojai}. We hypothesize that by enforcing consistency loss, we make the network less dependent on particular features (spatial, color related, etc.), which should nullify the effect of triggers, regardless of what specific dataset is used for computing and enforcing consistency.

%Without the label, the loss function is forced to reconcile the differences in predictions between perturbations of an image, and its end effect is to reduce the effect of the trigger.  
%Although this example outlined the case for point-triggers, feature-space triggers have a similar analog with other augmentations, such as color modifications.

\section{Experimental Setup} \label{sec:exp_setup}
We designed experiments to test our proposed UDA approach and answer three fundamental questions related to trojan mitigation:

\begin{enumerate}
    \item \textit{Algorithm Sensitivity}: How sensitive are sanitization algorithms to: 1) different types of trojans, 2) model architectures, and 3) the amount of supervised data made available for sanitization?
    \item \textit{Degradation of Non-Trojaned models}: How do sanitization algorithms affect models without trojans?
    \item \textit{Source of Unsupervised Data}: Can sanitization performance be increased by using unsupervised data? What characteristics of unsupervised data work best for sanitization?
\end{enumerate}

Our experimental matrix consists of two model architectures, two trigger types, two trojan behaviors, and two alternate datasets for unsupervised learning. The target task for our experiment is classification of the CIFAR-10 dataset~\cite{krizhevsky2009learning}. We generate different samplings of CIFAR-10 train and test sets, including samplings of which images to poison, and consider three different sizes of validation datasets for sanitizing the models. Details are provided in the following sections.

\subsection{Trojaned Dataset Configurations}
%To measure sensitivity to types of trojans, 
Our experiments uses combinations of two triggers and two trojan behaviors inserted into the CIFAR-10 dataset:
\begin{enumerate}
    \item Gotham Instagram filter applied to all classes
    \item Gotham Instagram filter applied to one class
    \item Reverse lambda pattern placed at the upper left corner applied to all classes
    \item Reverse lambda pattern placed at the upper left corner applied to one class
\end{enumerate}
The trojan behavior is configured such that when the corresponding trigger is present, the network learns to predict the next class, according to the CIFAR-10 dataset class enumeration \cite{krizhevsky2009learning}.  This variation in trigger types and trojan behaviors allows us to explore the difference in trojan mitigation performance for both global triggers (Instagram filter) and point triggers (reverse lambda pattern).  \emph{Global} refers to the fact that the trigger is applied across the entire image, whereas point triggers are localized to a certain region of the image.  

For each trojan and model architecture combination, we generate 5 Monte Carlo variants of trojaned models, with random subsets of triggered data chosen by different random seeds.  The datasets are combined into experiment configurations that specify the data points each model is trained with.  We utilize the TrojAI software framework \cite{karra2020trojai} to train the models, employing the standard approach of embedding trojans into models through data poisoning \cite{gu2017badnets} \footnote{All experimental configurations and training code will be released at \url{https://github.com/sanitais/}}. 

\subsection{Training the Trojaned Models}
\begin{figure}
    \centering
    \subfigure[]{\includegraphics[width=0.25\columnwidth]{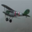}} 
    \subfigure[]{\includegraphics[width=0.25\columnwidth]{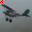}} 
    \subfigure[]{\includegraphics[width=0.25\columnwidth]{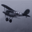}}
    \caption{(a) Original Image (b) Image with Reverse Lambda Trigger (c) Image with Instagram Gotham filter }
    \label{fig:trigger_examples}
\end{figure}

The trojaned models are generated by poisoning $20\%$ of the training data with triggers described above.  We chose $20\%$ to ensure that the model maintains good performance on clean data while also being responsive to the trigger.  Triggered image examples are shown in Fig.~\ref{fig:trigger_examples}.

To measure sensitivity to DNN model architecture, we conduct all experiments with both the VGG16 and WideResNet-28x10 network architectures~\cite{simonyan2014very, zagoruyko2016wide}.  The models are trained with the PyTorch framework \cite{pytorch} for 300 epochs, using stochastic gradient descent with a momentum of $0.9$, and a weight decay of $10^{-4}$.  The learning rate is set to $0.0025$ with a scaling factor $\lambda$ defined by Eq.~\ref{eq:lr_sched}. We include data normalization and randomized flips as part of the data pipeline. We exclude randomized cropping from the preprocessing pipeline due to its interference with the reverse-lambda trigger. This configuration is chosen because it is a common method of training the chosen model architectures for the CIFAR-10 dataset and provides a good balance between training time and performance.

\begin{equation}
    \lambda = \begin{cases}
    1 &\frac{\text{epoch}}{300} \leq 0.25\\
    1 - \frac{\left(\frac{\text{epoch}}{300}-0.25\right)}{0.65} \times 0.99 & 0.25 \leq \frac{\text{epoch}}{300} \leq 0.9\\
    $0.01$ & \frac{\text{epoch}}{300} > 0.9
    \end{cases}
    \label{eq:lr_sched}
\end{equation}

\subsection{Evaluation Metrics} \label{sec:eval_metrics}
We define three metrics for evaluation that capture degradation of the model on clean data and effectiveness at removing the response to the trigger.  Denote $y_{true}^{i}$ to be the correct label for input $x_i$, and $y_{trig}^{i}$ to be the label the network is configured to predict if the trigger is present.  Let $\mathcal{D}_{clean}$ represent a subset of the dataset $\mathcal{D}$ which only contains clean examples, and $\mathcal{D}_{trig}$ represents a subset of $\mathcal{D}$ that only contains triggered examples.  Additionally, define $|\mathcal{D}|$ to be the number of data points in dataset $\mathcal{D}$. In our test sets, we configure data points in $\mathcal{D}_{trig}$ to be $$y_{trig}^{i} = (y_{true}^{i} + 1) \mod  C$$ where $C=10$ is the number of classes.  Then, for a given model $M$, where $M(x)$ denotes the model output given input $x$, we define: 

\begin{equation}\label{eq:clean_acc}
    ACC_{clean} = \frac{\sum_{x_i \in D_{clean}} \left[M(x_i)=y_{true}^{i}\right]} {|\mathcal{D}_{clean}|}
\end{equation}

\begin{equation}\label{eq:fullrestore_acc}
    ACC_{fullrestore} = \frac{\sum_{x_i \in D_{trig}} \left[M(x_i)=y_{true}^{i}\right]} {|\mathcal{D}_{trig}|}
\end{equation}

\begin{equation}\label{eq:erase_acc}
    ACC_{erase} = \frac{\sum_{x_i \in D_{trig}} \left[M(x_i)=y_{trig}^{i}\right]} {|\mathcal{D}_{trig}|}
\end{equation}

Clean data accuracy, $ACC_{clean}$, represents the accuracy of the sanitized model on a held-out test set with no triggers.  Predictions are considered correct if the model predicts the correct label.  Full restore triggered data accuracy, $ACC_{fullrestore}$, represents the accuracy of the sanitized model on triggered data, where inference is considered correct if the model predicts the true label on triggered data.  A correct prediction indicates that the trigger has been nullified. Trigger erase accuracy, $ACC_{erase}$, represents the accuracy of the sanitized model on triggered data, where inference is considered correct if the model predicts the triggered label on triggered data.  A correct prediction indicates that the trigger is still in effect. A good sanitation algorithm will have a high clean data accuracy, a high full-restore triggered accuracy, and a low trigger-erase accuracy.  Note that full-restore accuracy and erase accuracy are not strict complements of each other.

% \subsection{Effect of Additional Data}
% To evaluate whether additional datasets can be helpful in sanitization, we experiment with CINIC-10~\cite{darlow2018cinic} and ImageNet~\cite{deng2009imagenet} datasets.  We use the these datasets as unsupervised datasets for UDA.  CINIC-10 is a drop-in replacement dataset for CIFAR-10 that contains different images of the same classes as CIFAR-10. ImageNet is a much larger dataset which contains many more data points and classes of data than CIFAR-10 and CINIC-10, and less overlap to CIFAR-10 than CINIC-10. Due to the size of ImageNet, we use a subset created using Numpy's \cite{numpy} random sampler with a seed of $1234$ to sample $10\%$ of the data, without applying class stratification.  This simulates a realistic scenario for trojan mitigation, where potentially unrelated data exists and is available for trojan mitigation. By noting that CINIC-10 and ImageNet are related to the original task dataset, CIFAR-10, in different ways, we can also answer the question of which type of additional data is best for improving sanitization performance, helping to address Q4 above.

\subsection{Sanitizing Models}
We compare our proposed approach with the latest state-of-the-art in trojan mitigation techniques, including fine-tuning, bridge mode connectivity (BMC), Neural Attention Distillation (NAD), Maxup and Cutmix augmentation \cite{gong2020maxup, yun2019cutmix}, and our own version of fine-pruning based on learning rate rewinding~\cite{renda2020comparing}, which we refer to as Learning-Rate rewinding and Compression, or LRComp.  Importantly, we note that every sanitization algorithm we evaluate is configured with the recommended hyperparameters outlined in the respective publication, to the extent possible. We configure UDA according to the default settings provided under the \textit{original} UDA use case, which is \emph{not} trojan mitigation.

For fine-tuning, we take advantage of the NAD codebase\footnote{\url{https://github.com/bboylyg/NAD}}, and accomplish fine-tuning by setting the $\beta$ parameter to zero, removing the NAD loss term. We found that the learning rate (LR) schedule from \cite{li2021neural} would often fail to clean the our models, and in some cases cause them to revert to random performance. With minimal testing of alternate LR schedules, we trained our models with the following settings: For the Gotham trigger, we set an initial LR of $0.1$ for our WideResNet architecture, and $0.001$ for our VGG16 architecture, and then multiply that rate by $0.1$ every two epochs (as done in \cite{li2021neural}). For the reverse-lambda trigger and both architectures, we use an LR of $0.02$ for epochs 1-3, $0.01$ for epochs 4-6, and $0.001$ for epochs 7-10. We train for a total of 10 epochs for each model.

For BMC, we use the default hyperparameters and training methodology outlined in \citet{zhao2020bridging}.  More specifically, we train for $600$ epochs with an initial LR of $0.015$, an LR schedule defined by Eq.~\ref{eq:lr_sched}, a weight decay of $5 \times 10^{-4}$, and a Bezier curve for the bridge with three control points.  Because BMC requires two models for the algorithm, we connect two triggered models with the same performance characteristics, but trained with different subsets of triggered data.  The exact details of which models were used for the experiments are provided in the open-source experimental configuration. The results reported for BMC correspond to the point along the curve which corresponds to $t=0.1$ for the VGG16 model, and $t=0.2$ for the Wide ResNet model, in accordance with the methodology reported in ~\citet{zhao2020bridging}.
%We emphasize that this is a generous and inflated way to measure the BMC algorithm's performance, as the triggered data is not available in a realistic scenario in order to determine the optimal value of $t$. However, we chose to compare with BMC at its best to make for a more convincing argument.

For NAD, we followed the procedure outlined by \citet{li2021neural}, with the same modifications as used in the fine-tuning method described above. We obtain the teacher model through the fine-tuning process previously described, then train using the same code and hyperparameters used in the fine-tuning step, but with the NAD loss parameter, $\beta$, set to $5000$, to obtain the NAD-sanitized model. Attention was computed using the $\mathcal{A}_{\text{sum}}^2$ attention map at the end of the convolution layers of our architectures, as done in \cite{li2021neural}.  The results reported correspond to the training epoch for which the model exhibits the highest $ACC_{clean}$. 

For LRComp, we fine-tune each model for $50$ epochs while decaying the initial learning rate of $0.001$ by a factor of $0.5$ at every epoch. Then, every five epochs, we remove (zero-out) the lowest magnitude 20\% of the currently active weights and reset the learning rate back to $0.001$.

For UDA, we train the networks for $200$ epochs, with the SGD optimizer set to a learning rate of $0.01$, Nestrov momentum of $0.9$ and a weight decay of $1 \times 10^{-4}$.  We also utilize a cosine annealing learning rate scheduler configured with a minimum learning rate of $1.2 \times 10^{-4}$.  These settings for training come from a reference implementation of UDA\footnote{\url{https://github.com/lantgabor/Unsupervised-Data-Augmentation-PyTorch}}, which we utilized in our experiments.  Four classes of UDA experiments are conducted: 1) UDA with no additional unsupervised data, 2) UDA augmented with in-class data from another source (CINIC-10) \cite{darlow2018cinic}, 3) UDA augmented with unsupervised random-class data from another source (ImageNet) \cite{deng2009imagenet}, and 4) UDA with no supervised data.  During training, we store the best model as measured by the accuracy on clean data, and use that model to compute the triggered data metrics, mentioned previously.  For computing the UDA consistency loss with unsupervised data, we use RandAugment \cite{cubuk2019randaugment} to produce randomized perturbations of the unsupervised input label.

Finally, to determine whether the consistency constraint imposed by UDA is a driver of santization performance, or whether complex data augmentations are sufficient, we test the performance of fine-tuning trojaned models with complex and aggressive data augmentations and the MaxUp loss function, which optimizes for the worst-case loss over augmented data~\cite{gong2020maxup}.  We combine this with CutMix augmentation, which combines random snippets of images from a configurable $m$ classes to confuse classifiers~\cite{yun2019cutmix}.  The combination of MaxUp and CutMix was shown to achieve the best performance for top1 and top5 accuracies on the validation set of ImageNet for a wide variety of model architectures  For these experiments, we train with a learning rate of $0.001$ for $200$ epochs using the SGD optimizer.  CutMix was configured with $m=4$, the same value which was used in the ImageNet experiments referenced.

\section{Results}
\label{sec:results}

\subsection{Algorithm Sensitivity}
We measure algorithm sensitivity to: 1) the amount of supervised data made available to the sanitization algorithm, 2) the type of trigger, 3) the type of trojan, and 4) the model architecture. We provide three different quantities of clean CIFAR-10 data to all sanitization algorithms: $5\%$, $10\%$, and $20\%$.  The four trigger-trojan configurations described above, combined with the two model architectures and five Monte-Carlo simulations per configuration, yields $120$ models to be sanitized for each of the six algorithms that we test.

\begin{figure*}[t!]
    % \subfloat{\includegraphics[width=\columnwidth]{figures/clean_test_acc.png}} 
    % \subfloat{\includegraphics[width=\columnwidth]{figures/full_restore.png}}\\
    % \subfloat{\includegraphics[width=\columnwidth]{figures/erase.png}}
    % \subfloat{\includegraphics[width=\columnwidth]{figures/clean_model_affect.png}}
    \includegraphics[width=\columnwidth]{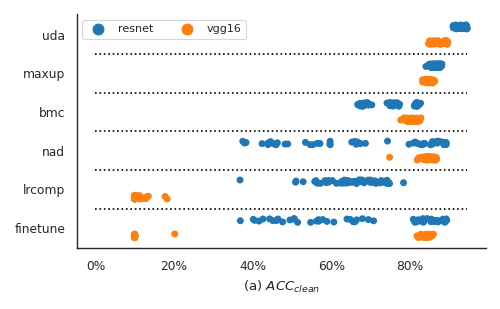}
    \includegraphics[width=\columnwidth]{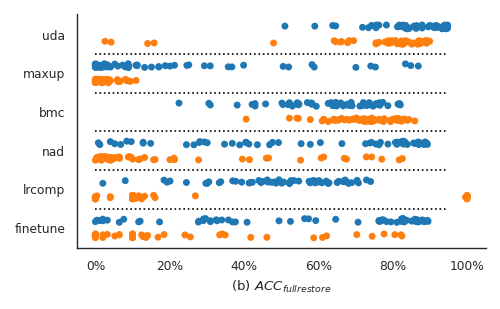}\\
    \includegraphics[width=\columnwidth]{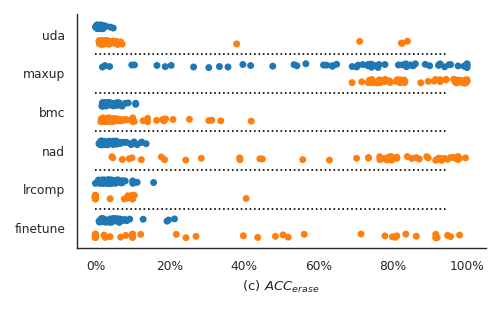}
    \includegraphics[width=\columnwidth]{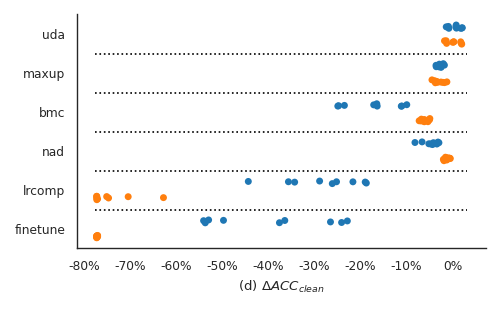}
    \caption{Performance after sanitization of ResNet and VGG16 models.  The $x$-axis label indicates the metric being measured, as previously defined in Section~\ref{sec:eval_metrics}. The plot was created by sweeping over different trojan configurations and amounts of supervised data, for both VGG16 and WideResNet-28x10 model architectures.}
    \label{fig:results}
\end{figure*}

%The results of sanitization for the tested configurations is shown in Fig.~\ref{fig:results}.  
Fig.~\ref{fig:results} (a), (b), and (c) display the values of the metrics defined in (\ref{eq:clean_acc}), (\ref{eq:fullrestore_acc}), and (\ref{eq:erase_acc}), respectively, of the various algorithms on the CIFAR-10 dataset for all trojan configurations, training data percentages, and model architectures.

The UDA results shown are with the configuration that included the CINIC-10 dataset to compute the unsupervised consistency loss.  Additional configurations of unsupervised datasets applied to UDA are compared and described in section \ref{sec:unsup-data}.  

Fig.~\ref{fig:results}(a), which displays the clean data accuracy $ACC_{clean}$, indicates that UDA outperforms all other compared algorithms for this metric.  It preserves the clean data performance across all compared model architectures, supervised data percentages, and trojan configurations.  Fig.~\ref{fig:results}(b), which measures trojan nullification, indicates that UDA generally performs better than the other tested algorithms. There are still cases where the algorithm failed to sanitize the network, as indicated by $ACC_{fullrestore}$ being low and corresponding examples of $ACC_{erase}$ being high.  Examining these cases in detail, we discovered that these results stemmed from the configuration where all classes were poisoned with the reverse lambda trigger and embedded into the VGG16 architecture. All other algorithms had similar difficulties with this configuration, except for BMC. We believe this merits further investigation, but for now leave as future work. On average however, the trends indicate UDA to still performs favorably compared to other algorithms across all performance metrics. Finally, Fig.~\ref{fig:results}(c) shows the effectiveness of the algorithm to erase the trigger. As noted previously, a lower value of $ACC_{erase}$ is desirable. UDA produces results closest to $0\%$.  The combination of Fig.~\ref{fig:results}(b) and (c) indicate that UDA generally performs best in removing trojans.

%The general trends indicate that UDA performs better than the other algorithms for all three metrics across model architectures, supervised data percentages, and trojan configurations. We observe that UDA outperformed the other sanitization algorithms for clean data accuracy and was less sensitive to the specific trojan configuration or amount of supervised training data.

\subsection{Degradation of Non-Trojaned models}
%\textcolor{red}{plot on how performance varies across the different trojans and data percentage - provides more detail, if you have space!?}

We test the effect of the sanitation algorithms on clean (non-trojaned) models.  In these experiments, we run a non-trojaned model through a sanitization algorithm, and measure the difference between $ACC_{clean}$ of the non-trojaned model processed by the algorithm, and the model before it was processed.  This measures any degradation in performance caused by the sanitization algorithm.  Here, the algorithms are configured in the exact same manner as above.  The difference in performance, denoted by $\Delta ACC_{clean}$ is shown in Fig~\ref{fig:results}(d).  The results show that UDA is the least detrimental amongst all algorithms for the tested configurations.

%The UDA results shown are again with the configuration that used the CINIC-10 dataset for computing the unsupervised consistency loss. 

%Additional performance results, categorized by the trojan type and the amount of data used for sanitization are shown in Appendix A.

\subsection{Source of Unsupervised Data}
\label{sec:unsup-data}
To evaluate whether additional datasets can be helpful in sanitization, we experiment with CINIC-10~\cite{darlow2018cinic} and ImageNet~\cite{deng2009imagenet} datasets.  We use the these datasets as unsupervised datasets for UDA.  CINIC-10 is a drop-in replacement dataset for CIFAR-10 that contains different images of the same classes as CIFAR-10. ImageNet is a much larger dataset which contains many more data points and classes than CIFAR-10 or CINIC-10, and has less overlap with CIFAR-10 than CINIC-10 does. Due to the size of ImageNet, we randomly select a $10\%$ subset without applying class stratification.  This simulates a realistic scenario for trojan mitigation, where potentially unrelated data exists and is available for trojan mitigation. Because CINIC-10 and ImageNet are related to the original task dataset, CIFAR-10, in different ways, we can also investigate the question of the characteristics of additional data that are best for improving sanitization performance.

\begin{figure}[t!]
    % \subfloat{\includegraphics[width=\columnwidth]{figures/uda_clean.png}} \\
    % \subfloat{\includegraphics[width=\columnwidth]{figures/uda_fullrestore.png}}\\
    % \subfloat{\includegraphics[width=\columnwidth]{figures/uda_erase.png}}
    \includegraphics[width=\columnwidth]{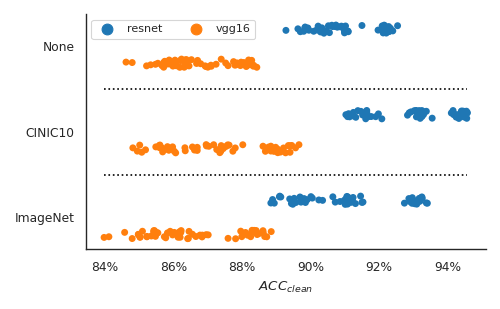} \\
    \includegraphics[width=\columnwidth]{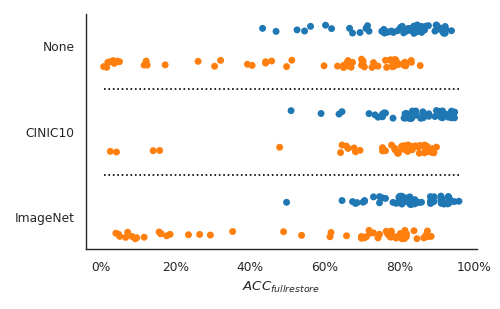}\\
    \includegraphics[width=\columnwidth]{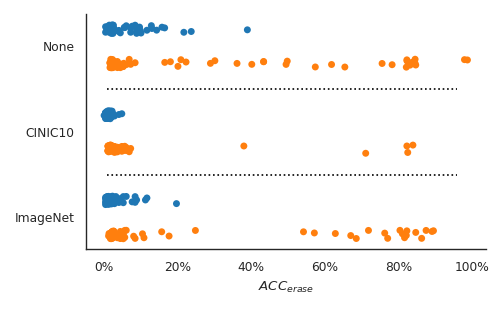}
    \caption{The performance of UDA after sanitization with different additional data sources on triggered data for the three metrics defined in Section~\ref{sec:eval_metrics}, across trigger-trojan configurations, model architectures, and supervised data percentages.}
    \label{fig:uda_results}
\end{figure}

The results are shown in Fig.~\ref{fig:uda_results}, aggregated across the four trigger-trojan pairs and supervised data percentages for both model architectures.  In these figures, the x-axis label defines the metric being measured, and the y-axis represents the additional dataset used for the UDA consistency loss (\ref{eq:uda_unsup}).  None indicates that no additional unsupervised data was used.

Figure~\ref{fig:uda_results} indicates that augmenting with the CINIC-10 dataset provides the greatest gain in performance.  This is intuitive, since the CINIC-10 dataset can be considered ``in-domain'' with CIFAR-10, or in other words, data from both datasets come from the same distribution.  The results indicate that ImageNet also provides gains, but they are not as pronounced as those coming from the used of CINIC-10.  Quantitatively, across all configurations, on average, the in-domain dataset (CIFAR-10) provides a $1.4\%$ increase in $ACC_{clean}$, $15.1\%$ increase in $ACC_{fullrestore}$, and $66.3\%$ \emph{decrease} in $ACC_{erase}$ when compared to using ImageNet for UDA.  This indicates that the in-domain data is preferred to UDA.  However, we also note that the UDA based approach with out-of-domain unsupervised data for both $ACC_{clean}$ and $ACC_{fullrestore}$ still outperforms other compared algorithms, and is comparable for the $ACC_{erase}$ metric.

An additional test was conducted to measure the performance of UDA with no labeled CIFAR-10 data, and only use unlabeled CINIC-10 and ImageNet data.  In these scenarios, UDA was not able to remove the trigger and performed poorly, indicating that a small percentage of supervised data is needed to bootstrap the trojan mitigation process.

\section{Discussion}
\label{sec:disc}

The results in Section~\ref{sec:results} are summarized by the following observations:
\begin{enumerate}
    \item Applying UDA with unlabeled data coming from a similar distribution as the original task significantly removes trojan effects from trained models with minimal negative effects. 
    \item The UDA algorithm is robust to multiple types of trojans, network architectures, and amounts of data used for sanitization.
    \item UDA is the least detrimental to clean models; other algorithms degrade performance at varying degrees. These results hold across multiple types of trojans and varying network architectures.
    \item Additional related data also helps sanitization performance in UDA, as shown by the increase in performance by using ImageNet for UDA.
    %, in Fig.~\ref{fig:uda_results}.  However, Fig.~\ref{fig:uda_results} (b) and (c) indicate that sanitization is not as robust across the tested configurations without in-domain unsupervised data.
\end{enumerate}

Table \ref{tab:results_summary} summarizes the average performance of the tested algorithms across all configurations. UDA compares favorably to all other algorithms for both the $ACC_{clean}$ and $ACC_{fullrestore}$ metrics.  However, BMC displays less variance in the $ACC_{fullrestore}$ metric.  The remainder of the algorithms fail to effectively remove the trojan properly. %, since trojan removal would be indicated by both high $ACC_{fullrestore}$ and low $ACC_{erase}$ values.

\begin{table}
\footnotesize
\begin{tabular}{c||c|c|c}
     & $ACC_{clean}$ & $ACC_{fullrestore}$ & $ACC_{erase}$ \\
    \hline
    UDA & $\mathbf{90.2 \pm 3.2}$ & $\mathbf{80.9 \pm 16.2}$ & $4.9 \pm 14.5$ \\ \hdashline
    BMC & $77.9 \pm 4.8$ & $67.4 \pm 12.4$ & $6.4 \pm 6.7$ \\ \hdashline
    NAD & $77.6 \pm 14.6$ & $37.9 \pm 33.3$ & $37.6 \pm 39.1$ \\ \hdashline
    \shortstack[c]{MaxUp \\ + \\CutMix} & \shortstack[c]{ \vphantom{MaxUp} \\ $85.8 \pm 1.1$ \\ \vphantom{CutMix}} & \shortstack[c]{ \vphantom{MaxUp} \\ $11.9 \pm 19.8$ \\ \vphantom{CutMix} } & \shortstack[c]{ \vphantom{MaxUp} \\ $76.8 \pm 24.0$ \\ \vphantom{CutMix} } \\ \hdashline
    LRComp & $37.8 \pm 27.7$ & $38.3 \pm 31.1$ & $\mathbf{5.0 \pm 5.1}$ \\ \hdashline
    FineTune & $58.0 \pm 31.4$ & $37.0 \pm 33.3$ & $17.1 \pm 27.5 $ 
\end{tabular}
\caption{Summary of results for all algorithms over all models, trojan configurations, and supervised data percentages.}
\label{tab:results_summary}
\end{table}

The difference in performance between UDA and MaxUp+CutMix indicates that the performance benefit gained by UDA is not due \textit{solely} to the randomized perturbations inherent in computing the unsupervised consistency loss, but also the fact that UDA is able to leverage additional data sources to improve performance.

We additionally note that because BMC builds a bridge-mode, there are an infinite amount of models to choose from along the curve for sanitization performance evaluation. As mentioned previously, we choose the points along the curve for each model architecture as recommended by the authors of the publication. However, it is likely that there exist other values of $t$ for which clean performance is better and the trojan effect is better mitigated. When triggered data is available and the trojan is known, one can attempt to find an optimal $t$ that maximizes $ACC_{clean}$ while minimizing $ACC_{erase}$, but it is not clear how one might choose $t$ in a realistic scenario where this information and data would be unavailable to the owner of the model. %While BMC has attractive qualities for trojan mitigation, unless a heuristic or other method of choosing a reliable $t$ value is developed, it will be an impractical technique. 

\section{Conclusion}
\label{sec:conclusion}
In this work, we have shown the efficacy of UDA in mitigating trojans for neural networks. The primary advantages of UDA over other methods are practical, in that: 1) the algorithm is robust to variants of triggers, models, and available data, 2) it can additionally leverage out-of-domain datasets to further boost sanitization performance, and 3) the generality of the UDA framework allows for the same algorithm to be applied across a variety of data modalities.  In our study, we found that a shortcoming of all of the current methods for trojan mitigation (including UDA) is that they require some minimum percentage of supervised data. Potential future work could address this via new consistency loss functions, newer algorithmic approaches to model sanitation, and stronger forms of augmentations such as those proposed by \citet{gong2020maxup}.
%Additionally, UDA is especially pertinent to trojan mitigation due to its ability to leverage unlabeled data, and its unifying framework that can handle multiple modalities of data, such as vision and text .  
% \section*{Software and Data}

% If a paper is accepted, we strongly encourage the publication of software and data with the
% camera-ready version of the paper whenever appropriate. This can be
% done by including a URL in the camera-ready copy. However, \textbf{do not}
% include URLs that reveal your institution or identity in your
% submission for review. Instead, provide an anonymous URL or upload
% the material as ``Supplementary Material'' into the CMT reviewing
% system. Note that reviewers are not required to look at this material
% when writing their review.

% % Acknowledgements should only appear in the accepted version.
% \section*{Acknowledgements}

% \textbf{Do not} include acknowledgements in the initial version of
% the paper submitted for blind review.

% If a paper is accepted, the final camera-ready version can (and
% probably should) include acknowledgements. In this case, please
% place such acknowledgements in an unnumbered section at the
% end of the paper. Typically, this will include thanks to reviewers
% who gave useful comments, to colleagues who contributed to the ideas,
% and to funding agencies and corporate sponsors that provided financial

\bibliography{aaai22}
\bibliographystyle{icml2022}

% %%%%%%%%%%%%%%%%%%%%%%%%%%%%%%%%%%%%%%%%%%%%%%%%%%%%%%%%%%%%%%%%%%%%%%%%%%%%%%%
% %%%%%%%%%%%%%%%%%%%%%%%%%%%%%%%%%%%%%%%%%%%%%%%%%%%%%%%%%%%%%%%%%%%%%%%%%%%%%%%
% % APPENDIX
% %%%%%%%%%%%%%%%%%%%%%%%%%%%%%%%%%%%%%%%%%%%%%%%%%%%%%%%%%%%%%%%%%%%%%%%%%%%%%%%
% %%%%%%%%%%%%%%%%%%%%%%%%%%%%%%%%%%%%%%%%%%%%%%%%%%%%%%%%%%%%%%%%%%%%%%%%%%%%%%%
% \newpage
% \appendix
% \onecolumn

% You can have as much text here as you want. The main body must be at most $8$ pages long.
% For the final version, one more page can be added.
% If you want, you can use an appendix like this one, even using the one-column format.
% %%%%%%%%%%%%%%%%%%%%%%%%%%%%%%%%%%%%%%%%%%%%%%%%%%%%%%%%%%%%%%%%%%%%%%%%%%%%%%%
% %%%%%%%%%%%%%%%%%%%%%%%%%%%%%%%%%%%%%%%%%%%%%%%%%%%%%%%%%%%%%%%%%%%%%%%%%%%%%%%

\end{document}